\documentclass[conference]{IEEEtran}
\usepackage[T1]{fontenc}
\usepackage[utf8]{inputenc}
\usepackage{cite}
\usepackage{amsmath,amssymb}
\usepackage{url}
\usepackage{booktabs}
\usepackage{array}
\usepackage{enumitem}
\usepackage{tikz}
\usetikzlibrary{arrows.meta,positioning,calc,fit,backgrounds}
\usepackage[switch]{lineno}
\setlist[itemize]{leftmargin=*,nosep}
\IEEEoverridecommandlockouts
\begin{document}
\title{When Prediction Shapes the Future: A BlueSky Agenda for Reflexive Time Series Analysis

\thanks{Yiqun Zhang is the corresponding author.}
}

\author{
\IEEEauthorblockN{
Tao Xie\textsuperscript{1},
Yiqun Zhang\textsuperscript{2,*},
Yiu-ming Cheung\textsuperscript{3}
}
\IEEEauthorblockA{
\textit{$^{1}$School of Automation, Guangdong University of Technology}\\
\textit{$^{2}$School of Future Technology, South China University of Technology}\\
\textit{$^{3}$Department of Computer Science, Hong Kong Baptist University}\\
3123001085@mail2.gdut.edu.cn,\;
yqzhangzyq@gmail.com,\;
ymc@comp.hkbu.edu.hk
}
}
\maketitle


\begin{abstract}
Time-series models are increasingly deployed in operational systems (e.g., inventory management and clinical care) where predictions trigger interventions that alter future observations, producing records we term \emph{AI-shaped trajectories}. 
This sequence-prediction coupling deepens as such systems evolve into agents that autonomously predict, act, and remember across long horizons. 
Such \emph{AI-shaped trajectories} support immediate performance monitoring, but turn ambiguous once reused as passive data. 
An algorithm may then misread a prediction-prevented stockout as naturally stable demand, or an averted medical crisis as a false alarm. 
This ambiguity then propagates into everything built on the record, from retraining and agent memory to downstream predictive tasks. 
As a result, systems may internalize these intervention effects as natural phenomena, drifting further from the truth at every closed-loop cycle. 
In this Blue Sky paper, we advocate an agenda for reflexive time-series analysis that shifts the analytical object from the intervention-agnostic sequence to the prediction--action--observation loop. 
The agenda requires closed-loop provenance to track the operational context (e.g., predictions, policies, and actions), sequence identification to estimate how those actions affect future values under explicit causal assumptions, and closed-loop evaluation to separate intrinsic process prediction from policy-driven effects. 
This foundation is essential for integrating autonomous agents into such loops at scale, 
enabling the joint learning of intervention effects and intrinsic dynamics for robust temporal representations. 
\end{abstract}

\begin{IEEEkeywords}
Time series mining, performative prediction, closed-loop evaluation, representation learning, forecasting
\end{IEEEkeywords}

\section{Introduction}
Time series are predominantly conceptualized as passive records of an exogenous process. 
This assumption holds primarily for non-interventional settings, 
where model outputs do not feed back into the underlying data-generating mechanism. 
However, this premise becomes untenable when deployed predictive systems drive an operational policy~\cite{mandi2024decisionfocused,wilder2019melding}, 
as actions based on predictions can alter subsequent observations. 
Prior work on performative prediction~\cite{perdomo2020performative, zhao2025performative} shows that deployed predictions can influence the future target distributions. 
Moving beyond static distribution shifts, we focus on the sequential artifacts continuously inscribed by these interventions, a phenomenon we term the \emph{AI-shaped trajectory}. 
Once reused for model retraining, benchmarking, or agent memory, these \emph{AI-shaped trajectories} carry both the underlying process dynamics and the effects of the policy under which they were generated. 
As systems become increasingly autonomous~\cite{cheng2026agentic}, this reuse renders the pipeline self-referential, thereby conflating action effects with the intrinsic dynamics of the underlying process.

\begin{figure}[t]
\centering
\includegraphics[width=0.85\linewidth]{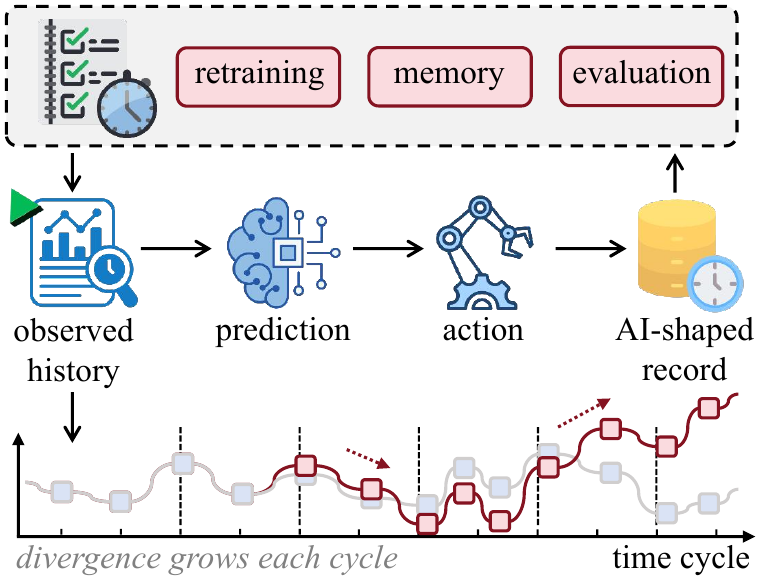}
\caption{Reflexive generation of an \emph{AI-shaped trajectory}. A forecast-driven action alters subsequent observations, which can re-enter later training, memory, or evaluation. Repeated feedback can progressively separate the observed trajectory from the underlying process dynamics.}
\label{fig:loop}
\end{figure}

In inventory control, a demand-forecasting model drives replenishment~\cite{makridakis2022m5accuracy,elmachtoub2022spo}. When a predicted stockout triggers early restocking, the shortage is prevented, and recorded sales remain flat. This observation is causally confounded: it might reflect intrinsically stable demand, or volatile demand successfully mitigated by the intervention. Clinical monitoring exhibits a similar pathology~\cite{kim2025monitoring}. 
A model flags a patient for deterioration, prompting preventive care that stabilizes the patient. 
Consequently, subsequent evaluations may misinterpret the initial alert as a false alarm. 
A similar closed loop arises in smart-grid demand response, where short-term load forecasts can determine dynamic electricity prices or load-shifting decisions, which can in turn influence subsequent consumption and make the realized load partly dependent on the forecasting-and-control policy~\cite{palaniyappan2025optimized}. Across these settings, the observational data records an outcome shaped by the intervention, thereby masking the underlying process dynamics.
This confounding becomes particularly critical in autonomous agents that forecast, act, and retrain on the records their own actions produce. 
Without external calibration mechanisms, the causal ambiguity is easily overlooked, causing the system to continuously internalize its own policy effects as inherent natural phenomena.


Standard action logs document interventions but exclude counterfactual outcomes~\cite{bica2020crn,melnychuk2022causaltransformer}. Consequently, the \emph{AI-shaped trajectory} entangles the intervention effect with the underlying process dynamics, rendering the two components difficult to disentangle from observed values alone~\cite{runge2023causalts}. This mixture spreads through routine downstream applications, effectively corrupting three fundamental phases: 1) \textit{Model retraining} on these trajectories injects past policy effects into the newly learned dynamics, exacerbating the structural bias with each successive cycle~\cite{adam2022error,taori2023datafeedback,wyllie2024fairness}; 2) When repurposed for \textit{performance evaluation}, benchmarking heavily rewards agreement with policy-constrained futures, mistaking self-fulfilling prophecies for genuine predictive capability~\cite{time2026benchmark}; 3) Utilizing these trajectories as \textit{agent memory} forces the system to internalize its own interventions as objective environmental truths~\cite{xiong2025memory}. 
All three phases operate on the flawed assumption of an undisturbed data-generating process. However, because the AI-shaped trajectory preserves only the final outcome and omits the causal link between action and effect, this assumption catastrophically fails, and standard predictive approaches inherently struggle to overcome this missing link. Therefore, extracting genuine process dynamics from such contaminated records necessitates a dedicated analytical framework.

In this Blue Sky paper, we propose an agenda for \emph{reflexive time-series analysis} that explicitly acknowledges the system's active role in shaping its own future data. This agenda shifts the primary unit of analysis from the intervention-agnostic sequence to the dynamic, closed loop of prediction, action, and observation. The agenda unfolds across three essential stages: 1) Closed-loop provenance traces and records the operational context, encompassing the deployed prediction, the governing policy, and the executed action; 2) Building upon this foundation, sequence identification integrates explicit causal assumptions to estimate how these interventions perturb future sequence values; 3) Closed-loop evaluation leverages these counterfactual estimates to disentangle intrinsic process predictability from policy-driven artifacts. As autonomous agents increasingly govern these operational loops at scale, this three-stage foundation provides the external calibration required to break the self-referential cycle. By jointly learning intervention effects and intrinsic dynamics, this agenda produces robust temporal representations that maintain their scientific validity even as systems repeatedly reuse them for retraining, benchmarking, and agentic memory.

The remainder of this paper is organized as follows: Section~\ref{sec:idea} formalizes the analytical object and examines its urgency under current deployment trends. Section~\ref{sec:oppchal} develops the three-stage agenda of loop provenance, sequential identification, and closed-loop evaluation. Section~\ref{sec:success} establishes criteria for methodological progress, and Section~\ref{sec:frontiers} extends this framework to broader data mining applications.

\section{The Reflexive Paradigm: From Passive Records to Active Loops}
\label{sec:idea}

Time-series mining has long prioritized predicting future sequences over analyzing the temporal records left behind by deployed models. 
An AI-shaped trajectory should therefore be treated not as an intervention-agnostic sequence, but as empirical evidence of the dynamic loop that produced it.
\emph{Reflexive time series analysis} therefore shifts the analytical unit from the intervention-agnostic sequence $X_{1:T}$ to the 
underlying prediction-action-observation mechanism.
Standard forecasting frameworks inherently lack the capacity to recover this mechanism, demanding new infrastructure built on three capabilities: closed-loop provenance, sequential identification, and rigorous evaluation.

The phenomenon of predictions reshaping their target environments is well documented across disciplines, spanning self-fulfilling prophecies, Goodhart's law, and performative prediction~\cite{king2023selffulfilling,perdomo2020performative}. The gap lies in how time-series mining formalizes the resulting sequence. 
Current practice discards this causal context: benchmarks expose recorded values while omitting the prediction, policy, action, operational context, and counterfactual uncertainty needed for deployment interpretation. Without this information, the record is difficult to interpret because standard evaluation relies on an assumption that deployment can violate. This assumption is causal rather than statistical: although a valid forecast should correlate with its target, evaluation typically assumes that the target itself is not altered by the forecast~\cite{perdomo2020performative,hardt2023performative}. Under operational policies, this assumption may fail, rendering outcomes policy-dependent and motivating a dedicated analytical formulation and evaluation infrastructure.

Two structural shifts necessitate this paradigm. First, the prediction-to-action pathway is increasingly designed rather than incidental. Decision-focused and control-oriented methods now train forecasts to drive actions rather than serve as terminal outputs~\cite{mandi2024decisionfocused,beichter2025decisionfocused,rawlings2017mpc}, foundation models lower barriers to deploying such forecasters across domains~\cite{ansari2024chronos,goswami2024moment}, and agentic systems make the loop persistent by executing actions and storing outcomes in memory~\cite{cheng2026agentic}. Consequently, the passive-record assumption fails at scale. Second, the empirical evidence of this feedback is broad, spanning model-induced distribution shifts~\cite{perdomo2020performative,kehrenberg2026performative}, bias amplification in data pipelines~\cite{wyllie2024fairness}, feedback in deployed recommenders and clinical risk scores~\cite{mansoury2020feedback,rosenthal2025dynamic}, and formal performativity in time-series forecasting~\cite{zhao2025performative,bhati2022performativets}. Although existing performativity-aware methods anticipate induced shifts to preserve accuracy, predictive accuracy alone cannot establish how the underlying loop should be recorded, identified, and evaluated. Without this causal context, downstream analyses may conflate intervention effects with underlying process dynamics. Maintaining explicit links among forecasts, policies, and actions preserves the evidence needed to examine this ambiguity under causal assumptions. Once these links are lost, the intervention context becomes unrecoverable from the sequence alone, which complicates subsequent reuse and motivates the proposed agenda. 


Formalizing this paradigm requires explicit definition of the loop's causal mechanics. In a minimal forecasting setting:
\begin{equation}
\widehat{Y}_t = f_\theta(X_{\le t}), \qquad
A_t = \pi(\widehat{Y}_t), \qquad
Y_{t+h}^{obs}=Y_{t+h}(A_t).
\end{equation}
Here, the model $f_\theta$ predicts $\widehat{Y}_t$ from history, policy $\pi$ triggers action $A_t$, and the outcome $Y_{t+h}^{obs}$ materializes strictly under this intervention. This continuous propagation distinguishes reflexive analysis from standard performative prediction~\cite{perdomo2020performative}. Rather than merely shifting a global data distribution, the action $A_{t}$ alters the underlying state space, dynamically distributing intervention effects across the entire subsequent trajectory. This creates a severe sequential identification problem, where naturally stable processes and action-suppressed volatile processes yield mathematically identical empirical observations. Since standard action logs cannot reveal counterfactual outcomes~\cite{bica2020crn,melnychuk2022causaltransformer}, resolving this ambiguity requires explicit structural assumptions, such as transition models or policy variation. Our agenda aims to formalize this causal disentanglement, ensuring models are evaluated against objective realities rather than artificial futures they helped construct.

\section{A Three-Tier Agenda for Reflexive Analysis}
\label{sec:oppchal}
 
Analyzing AI-shaped trajectories requires a strict dependency chain: recording the loop, identifying its causal effects, and evaluating predictive claims under closed-loop conditions. The objective is to build infrastructure for prediction--action--observation loops, rather than appending new metrics to passive benchmarks. 
 
\textbf{Upstream: recording the loop.}
To make forecast-mediated action part of the temporal record itself. A minimal provenance schema should store the value sequence along with the deployed model or model version, the forecast, the policy mapping, the action taken, the operational context, and the subsequent observed outcome~\cite{gebru2021datasheets,pushkarna2022datacards}:
\begin{equation}
(t, X_{\le t}, f_\theta, \widehat{Y}_t, \pi, A_t, H_t, Y_{t+h}^{obs}).
\end{equation}
In this schema, $H_t$ includes human overrides, constraints, or operational information available at decision time. In the inventory example, a complete record stores the sales series together with the forecast or stockout-risk estimate, the replenishment rule that used it, and the order placed under that rule. Such a record cannot identify action effects by itself; it can only make the intervention visible to downstream analysis. The challenge remains that forecasting, decision-making, and logging are often separated across software systems and organizational boundaries~\cite{sambasivan2021datacascades}. Because policies change and model versions update between cycles~\cite{mitchell2019modelcards}, analysts who do not record the links among model version, forecast, policy, action, and outcome observe only policy-dependent outcomes without the mechanisms that produced them. This raises the first open problem (O1): to define a minimal loop provenance standard that is expressive enough to support subsequent identification yet simple enough to be attached to deployed temporal data.

\begin{figure}[t]
\centering
\includegraphics[width=0.9\linewidth]{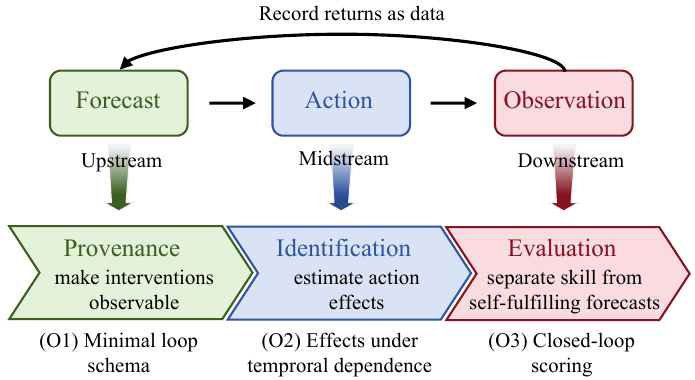}
\caption{A three-tier agenda for analyzing \emph{AI-shaped trajectories}. Provenance records the loop, identification estimates action effects, and evaluation separates predictive skill from self-fulfilling forecasts. Each tier builds on the one before it, and each ends in an open problem, numbered O1 to O3, marking a direction for future work.}
\label{fig:agenda}
\end{figure}

\textbf{Midstream: identifying action effects under temporal dependence.}
To identify action effects by modeling actions as components of the temporal state transition rather than as external metadata~\cite{ljung1999system,runge2023causalts}. This requires estimating how a sequence would evolve under alternative actions or no action~\cite{bica2020crn,melnychuk2022causaltransformer,hernan2020causal}. In the inventory example, this asks how the sales series would have evolved under a different order or no order.
In a smart-grid demand-response system, the analogous question is how the load trajectory would have evolved under a different price signal or without forecast-driven load shifting~\cite{palaniyappan2025optimized}.
Estimating such counterfactual evolution requires assumptions about the transition mechanism, deliberate policy variation, experimental interventions, simulators, or other sources of counterfactual information. Existing counterfactual and sequential decision methods provide foundations for action-dependent outcomes~\cite{swaminathan2015crm,levine2020offline}, but closed-loop temporal records add a propagation problem. The time-series difficulty is that an intervention propagates over time rather than affecting only an immediate instance. An action-conditioned model must represent both the direct effect of the action and its delayed effect on future states; otherwise, evaluation may reward models that fit immediate outcomes while missing feedback-driven state changes. The midstream open problem (O2) is to connect counterfactual estimation, state-transition modeling, and data conditions that make action effects estimable rather than merely logged.
 
\textbf{Downstream: evaluating under a closed loop.}
To design protocols that distinguish predictive skill from agreement with outcomes shaped by the forecast-policy system. In a closed loop, the recorded future may no longer be a neutral target, because it can be a policy-dependent outcome shaped by the forecast-policy system. Low error can have multiple interpretations. It may indicate that the model captured the intrinsic process dynamics, or that the system steered the process toward a state closer to the prediction. Conversely, an accurate risk forecast can appear inaccurate when the action it triggers prevents the event against which the forecast is later evaluated (Section~\ref{sec:idea}). Evaluation should therefore report the model, policy, action, and context that produced the observed target, not only the prediction error measured against that target. Performative prediction provides a starting point for analyzing model-induced distribution changes~\cite{perdomo2020performative,hardt2023performative}, and feedback detection identifies predictors that shape their own inputs~\cite{wager2014feedback}, but closed-loop time-series evaluation must additionally account for delayed temporal propagation and targets produced under policy intervention. The downstream open problem (O3) is to extend benchmark design beyond agreement with the recorded future~\cite{time2026benchmark, mu2026epicfbench}, defining metrics and protocols that separate skill at forecasting intrinsic process dynamics, policy effects, and self-fulfilling forecasts. 

\textbf{Across the tiers.}
The three tiers are interdependent: provenance identifies which interventions shaped the record, sequential assumptions support counterfactual estimation, and closed-loop evaluation distinguishes predictive skill from policy-shaped outcomes. This dependency becomes more important in agentic systems, where forecasts, actions, and stored outcomes are coupled and repeatedly reused for training, evaluation, or memory.


\section{Objectives of Reflexive Time Series Analysis}
\label{sec:success}

The previous section set out the tasks and open problems of the agenda. This section describes the end state those tasks aim at, that is, what a successful paradigm looks like in practice. Success here extends beyond improving passive forecasting accuracy to formalizing deployed records as artifacts of prediction--action--observation loops. This end state rests on four objectives. The first three carry the agenda over to the deployed setting, and the fourth is new.

\textbf{Routine loop provenance.} Datasets and benchmarks must encode the generative loop alongside the time series. Recording the model, forecast, policy, action, and context establishes the structural prerequisite for identifying intervention effects.

\textbf{Estimable action effects.} Counterfactual claims about how a trajectory evolves should rest on explicit assumptions, such as transition models or policy variation. Evaluations can then attribute performance gains to specific causal mechanisms rather than conflating them with intrinsic process dynamics.

\textbf{Disentangled evaluation.} Protocols should isolate genuine predictive skill from policy-dependent outcomes. Metrics should then distinguish whether reduced error reflects accurate intrinsic modeling or a self-fulfilling forecast. An averted risk should be read as a successful intervention rather than an apparent prediction failure.

\textbf{Auditable agentic memory.} As systems archive outcomes as reusable experience, their memory should remain auditable. An auditable memory preserves loop provenance and classifies each observation as an exogenous process, an intervention response, or an unresolved mixture.

Underlying these four objectives is a single structural shift, one that treats deployment as an integral part of the data-generating mechanism rather than an external disturbance.

\section{Generalizing Reflexivity: Beyond Forecasting to Agentic Systems}
\label{sec:frontiers}
 
The mechanism that produces AI-shaped trajectories also links this agenda to broader data mining frontiers, extending concerns from interpreting the temporal evidence behind individual predictions~\cite{xie2026anchormoe} to tracing how deployed models shape the records they subsequently consume~\cite{pagan2023feedbackloops}. Recommender systems shape exposure and then learn from the resulting clicks or purchases~\cite{mansoury2020feedback,klimashevskaia2024popularity}.
Models trained on synthetic or model-filtered data can amplify biases already present in the selected records~\cite{taori2023datafeedback,wyllie2024fairness}. Similarly, clinical systems updated after deployment create feedback once predictions influence care and subsequent outcomes~\cite{adam2022error,balcarcel2025feedback}. 
In each case, the record is produced after a model or policy has participated in the process it represents. Time series is distinct because interventions enter an evolving state where, unlike recommendation feedback localized to items or interactions, their effects propagate along the subsequent trajectory. 

This framing clarifies the role of future agentic temporal reasoning systems~\cite{chang2025reasoning}, whose lasting outputs are themselves stored as temporal records. An agent does not merely read such a record once; it stores it in memory, reusing it across many later decisions. Without loop provenance, each reuse feeds a policy-dependent outcome back into the next forecast as an independent exogenous observation. Consequently, the contamination described in the introduction grows cumulatively with each cycle, rather than remaining a fixed error. Conversely, when recorded and evaluated under the proposed agenda, these same records support reliable audit, adaptation, and reuse. The frontier is thus a specific mandate that time-series mining should learn to analyze records generated under systems that autonomously predict, act, and remember.


\section{Conclusion}
\label{sec:conclusion}
 

This paper establishes that the traditional passive-record assumption could be fundamentally invalidated when deployed forecasts dictate operational actions. Such coupling continuously generates policy-dependent outcomes that we conceptualize as \emph{AI-shaped trajectory}. Properly interpreting these contaminated records necessitates treating the dynamic prediction-action-observation loop as a primary analytical object, thereby enabling the separation of intervention effects from intrinsic process dynamics. Crucially, as temporal systems increasingly predict, act, and remember, standard accuracy metrics may fail to meaningfully evaluate a forecast against a future it actively helped to create. 
To overcome this methodological blind spot, we urge the time-series mining community to develop new data provenance standards, causal identification techniques, and closed-loop benchmarking protocols. 
Although deployed forecasting serves as the minimal instance of this phenomenon, the rapid transition toward autonomous temporal agents makes this reflexive agenda necessary.
Future investigations should therefore move beyond isolated prediction--action effects toward understanding how such effects accumulate through repeated retraining, memory reuse, and interactions among adaptive agents.
Ultimately, the illusion of passive observation is no longer tenable. The future of data mining lies in modeling how AI systems recursively shape their own empirical reality.

\bibliographystyle{IEEEtran}
\bibliography{references}
\end{document}